\documentclass{article}

\usepackage{float}
\usepackage[preprint]{corl_2026}
\usepackage{amsmath}
\usepackage{amssymb}
\usepackage{booktabs}
\usepackage{graphicx}
\usepackage{multirow}
\usepackage{array}
\usepackage{tabularx}
\usepackage{listings}
\usepackage{placeins}
\newcolumntype{Y}{>{\raggedright\arraybackslash}X}
\newcolumntype{Z}{>{\centering\arraybackslash}X}
\title{HRIBench: Benchmarking Interaction-Centric Human-Robot Collaboration}

\author{
Chang Liu\textsuperscript{1,*},
Jiawei Zhang\textsuperscript{1,*},
Tao Zhang\textsuperscript{1},
Ye Wang\textsuperscript{1},
Hongyu Zhou\textsuperscript{2},
Qin Jin\textsuperscript{1,\S} \\
\textsuperscript{1}Renmin University of China \\
\textsuperscript{2}Beijing Normal University \\
\textsuperscript{*}Equal contribution. \quad
\textsuperscript{\S}Corresponding author.
}

\begin{document}
\maketitle

\begin{abstract}
Current vision-language-action (VLA) benchmarks primarily evaluate isolated manipulation skills while leaving human-robot interaction structure largely unmodeled. However, real-world collaboration fundamentally requires coordination under shared agency, including intent understanding, temporal synchronization, protocol adherence, and safe interaction in dynamic environments. To address this gap, we introduce \textbf{HRIBench}, a diagnostic benchmark for intent-aware human-robot collaboration based on executable interaction scenarios. HRIBench represents collaborative tasks as structured scenario scripts that explicitly model agent roles, temporal dependencies, coordination constraints, and human behavior distributions. 
Building on this abstraction, HRIBench defines three representative interaction roles: \textit{Instructor}, \textit{Collaborator}, and \textit{Intruder}, covering intent communication, joint coordination, and robustness under human intervention. The benchmark contains 13 role-conditioned tasks with over 650 evaluation episodes generated from diverse interaction trajectories and scene variations. Beyond binary task success, HRIBench introduces interpretable interaction-centric metrics spanning synchronization, responsiveness, protocol compliance, and safety. 
We evaluate adapted policies based on GR00T, $\pi_{0.5}$, and ACT under a unified protocol. Results show that current foundation robot policies struggle substantially in collaborative settings despite strong manipulation ability, revealing major limitations in temporal coordination and intent-aware behavior. Fine-tuning on HRIBench consistently improves collaborative performance. In a real-world adaptation study, simulation data generated by HRIBench improves GR00T N1.5's physical-task success rate from $0.10$ to $0.43$, demonstrating the benchmark's value for advancing interaction-centric robot learning.

\end{abstract}


\section{Introduction}
\label{sec:introduction}
\begin{figure}[t]
    \centering
    \includegraphics[
        width=\linewidth,
        height=0.6\linewidth,
        trim=65pt 603pt 218pt 33pt,
        clip
    ]{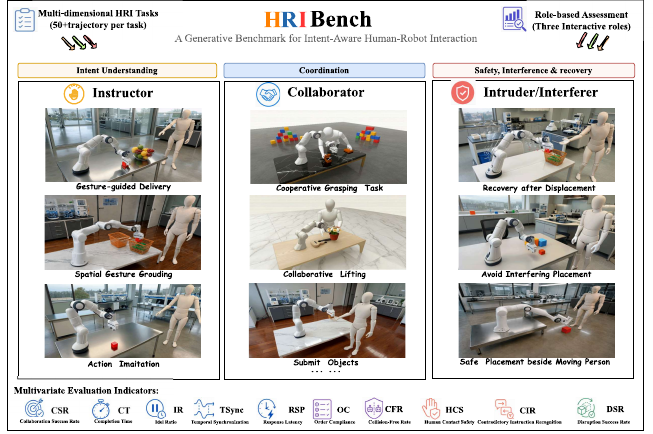}
    \vspace{-14pt}
    \caption{HRIBench evaluates robot policies in executable human-robot interaction scenarios. Each scenario explicitly specifies the human role, the robot objective, temporal coordination constraints, and the human motion distribution.}
    \label{fig:teaser}
\end{figure}
Vision-language-action (VLA) models have recently achieved rapid progress in language-conditioned robotic manipulation~\citep{brohan2023rt2,kim2024openvla,black2024pi0}. 
Large-scale robot policies can now generalize across diverse objects, instructions, and environments, leading to substantial advances in embodied foundation models. 
Existing benchmarks accordingly focus on manipulation-centric tasks such as pick-and-place, articulated object operation, and goal-conditioned execution. 
However, most current evaluations implicitly assume that robots operate alone.

Real-world robot deployment, by contrast, fundamentally involves interaction with humans. 
In collaborative settings, robots must not only manipulate objects, but also coordinate actions with people in shared environments. 
This requires reasoning about human intentions, temporal coordination, interaction protocols, and physical safety under shared agency~\citep{goodrich2007survey,sheridan2016hri,dragan2013legibility,huang2016anticipatory,strabala2013seamless,lasota2017safe,villani2018survey,argall2009survey}. 
A policy that succeeds in isolated manipulation may still fail to respond to a collaborator's reaching motion, violate coordination timing, or continue execution after disruptive human intervention. 
These failures expose a critical gap between manipulation competence and collaborative intelligence.

We argue that this gap arises because existing VLA benchmarks primarily evaluate \emph{manipulation outcomes}, while leaving the underlying \emph{interaction structure} between humans and robots largely unmodeled. 
Human-robot collaboration is fundamentally a coordination problem rather than a standalone manipulation problem. 
Effective collaboration depends not only on \emph{what} action is executed, but also on \emph{when}, \emph{why}, and \emph{in response to whom} the action is performed. 
This requires explicit modeling of interaction dependencies, including role assignments, temporal ordering, synchronization constraints, and human behavioral dynamics.

To address this limitation, we introduce \textbf{HRIBench}, a diagnostic benchmark for intent-aware human-robot collaboration. 
Rather than treating collaboration as isolated task execution, HRIBench represents interaction as executable coordination structure. 
Specifically, we introduce \emph{scenario scripts}, a structured and executable representation of collaborative interaction episodes. 
Each script explicitly specifies participating agents, interaction stages, temporal dependencies, coordination constraints, protocol ordering, and human motion descriptions. 
This abstraction provides an interpretable semantic layer connecting high-level interaction requirements to executable simulation environments, reward functions, and evaluation metrics.

A key design principle of HRIBench is to organize collaborative behavior around \emph{interaction roles}. 
We identify three fundamental role patterns of human behavior frequently encountered in human-robot interaction: \textit{Instructor}, \textit{Collaborator}, and \textit{Intruder}. 
Instructor scenarios evaluate whether robots can infer intended actions from human demonstrations or gestures. 
Collaborator scenarios require synchronized multi-agent coordination such as object handover or joint manipulation. 
Intruder scenarios test robustness under disruptive human intervention, requiring policies to maintain safety and adapt execution online. 
Importantly, these categories are not merely task groupings; they represent distinct coordination structures with different intent, timing, and safety requirements. Based on this framework, HRIBench instantiates 13 role-conditioned tasks with over 650 executable evaluation episodes generated from diverse interaction trajectories and scene variations. 
Rather than maximizing benchmark scale, we prioritize interaction completeness, interpretability, and diagnostic value. 
This design enables failures to be attributed to specific coordination deficiencies rather than hidden dataset heterogeneity.

Beyond binary task completion, HRIBench introduces an interaction-centric evaluation protocol spanning synchronization, responsiveness, protocol compliance, and safety. 
This allows collaborative failures to be decomposed into interpretable categories: a policy may fail because it misunderstands human intent, acts at an inappropriate time, violates coordination order, or enters unsafe proximity. 
Such disentangled evaluation is critical for understanding the limitations of current embodied foundation models in collaborative settings.
We evaluate representative robot policies adapted from modern VLA and imitation-learning backbones, including GR00T N1.5~\cite{groot2025}, $\pi_{0.5}$~\cite{physicalintelligence2025pi05}, and ACT~\cite{zhao2023act}. 
Experimental results show that although these models exhibit strong manipulation capabilities, they struggle substantially in collaborative interaction scenarios, particularly under temporal coordination and human perturbation: the best CSR remains $0.533$ on Instructor tasks and $0.500$ on Collaborator tasks, but drops to only $0.100$ on Intruder tasks. 
At the same time, HRIBench data improves real-world adaptation of GR00T N1.5, increasing the average physical-task success rate from $0.10$ with Real-only post-training to $0.43$ with Sim+Real post-training. 
These results demonstrate the benchmark's effectiveness for diagnosing interaction-centric failures and providing structured data for improving human--robot collaboration policies.

In summary, the main contributions of this paper are as follows:  (1) We introduce \textbf{HRIBench}, a diagnostic benchmark for human-robot collaboration that reformulates interaction evaluation as executable coordination modeling. HRIBench contains 13 role-conditioned tasks spanning instructor-guided execution, collaborative coordination, and robustness under human intervention. (2) We propose \textbf{scenario scripts}, a structured executable representation of collaborative interaction that explicitly models agent roles, temporal dependencies, coordination constraints, and human behavior distributions. (3) We establish an \textbf{interaction-centric evaluation protocol} for embodied policies, including synchronization, responsiveness, protocol compliance, and safety, enabling interpretable analysis of collaborative failure modes across modern robot foundation models.

\section{Related Work}
\label{sec:related_work}

\paragraph{Robot manipulation and VLA benchmarks.}
Robot learning benchmarks have made object manipulation measurable and comparable~\cite{li2023behavior,nasiriany2024robocasa}. RLBench~\citep{rlbench}, CALVIN~\citep{mees2022calvin}, ManiSkill~\citep{mu2021maniskill}, and Meta-World~\citep{metaworld} cover multi-task manipulation, language-conditioned execution, articulated-object interaction, and long-horizon control. In parallel, VLA policies such as RT-2~\citep{brohan2023rt2}, OpenVLA~\citep{kim2024openvla}, $\pi_0$~\citep{black2024pi0}, $\pi_{0.5}$~\citep{physicalintelligence2025pi05}, and GR00T~\citep{groot2025} have improved generalization across objects, scenes, and instructions. These systems are an important foundation for HRIBench: without manipulation competence, collaboration cannot succeed.The limitation is that most of these evaluations still end at an object-centric goal. A robot is usually judged by whether it opens a drawer, places an object, or follows a command, while the human remains outside the execution loop~\cite{shi2025hribench}. In real collaboration, however, the right action depends on a person's cue, timing, movement, or interference~\cite{shi2025hribench,jung2026semiac}. HRIBench therefore keeps the policy-level manipulation setting, but makes the human role part of the task definition rather than an external annotation.

\paragraph{Human-aware embodied evaluation.}
HRI research has long emphasized that collaboration depends on intent communication, legible behavior, temporal coordination, handover timing, and safe shared-space operation~\citep{argall2009survey,sauppe2014deictics,dragan2013legibility,huang2016anticipatory,strabala2013seamless,lasota2017safe,jung2026semiac,bo2026fast,wang2026anticipating,bejarano2026perception,hoffman2007effects,cakmak2011using}. Recent embodied benchmarks move in this direction by introducing humans, humanoid agents, or multi-agent task structure~\cite{wang2024demonstrating}. Habitat 3.0~\citep{habitat3} and PARTNR~\citep{chang2024partnr} study co-present agents and collaborative planning in household environments, while HandoverSim~\citep{handoversim} focuses on close-proximity handover. HRIBench is complementary to these efforts. It does not aim to replace navigation-scale embodied planning benchmarks or specialized handover simulators. Instead, it asks a narrower but policy-critical question: when a manipulation policy is placed in an executable interaction contract, can it infer the human role, act at the right time, obey the protocol, and recover safely when the human changes the scene~\cite{wang2026anticipating,jung2026semiac,baraglia2016initiative}?

\paragraph{Generative simulation.}
Generative simulation methods use language models or procedural pipelines to create robot tasks, scenes, rewards, and demonstrations~\citep{wang2023robogen,ma2024eureka,zhang2023large,nasiriany2024robocasa}. Human motion generation further provides diverse motion priors~\citep{guo2022humanml3d,petrovich2022temos,tevet2023mdm,guo2024momask,batool2026humandiffusion,zhang2024motiondiffuse,zhang2023generating}. These tools are useful for scaling environment construction, but a benchmark needs more than plausible generated content: episodes must be executable, reproducible, and tied to unambiguous metrics. HRIBench uses typed scripts, labeled human motions, success predicates, and simulation filtering to turn generated scenarios into standardized interaction episodes.

\section{Benchmark Design}
\label{sec:benchmark_design}

HRIBench is a diagnostic benchmark for evaluating intent-aware human--robot collaboration. 
It is organized around three interaction roles and contains 13 tasks with more than 50 trajectory variants per task, yielding over 650 executable evaluation episodes. 
Unlike conventional manipulation benchmarks that treat humans as passive scene context, HRIBench explicitly models the human as part of the task specification. 
Each role determines what information the human provides, how the human moves during execution, and which collaborative failure modes are exposed. 
Additional task details and the user study are provided in Appendix B and Appendix E.

Table~\ref{tab:benchmark_comparison} summarizes the diagnostic coverage of HRIBench relative to representative benchmarks. 
The goal is not to provide a general ranking, since existing embodied benchmarks target complementary capabilities such as large-scale manipulation or embodied planning. 
Instead, HRIBench focuses on interaction-centric evaluation by jointly supporting role-conditioned scripts, policy-level manipulation under dynamic human motion, perturbation and recovery protocols, interaction-specific metrics, and physical adaptation studies within a unified framework.

\begin{table}[t]
\centering
\small
\resizebox{\linewidth}{!}{
\begin{tabular}{l|ccccc}
\toprule
Benchmark & Role Script & Human Manipulation & Perturbation & Interaction Metrics & Physical Adaptation \\
\midrule
RLBench~\cite{rlbench} / CALVIN~\cite{mees2022calvin} & $\times$ & $\times$ & $\times$ & $\times$ & $\times$ \\
ManiSkill~\cite{mu2021maniskill} / Meta-World ~\cite{metaworld}& $\times$ & $\times$ & $\times$ & $\times$ & $\times$ \\
Habitat 3.0~\cite{habitat3} / PARTNR~\cite{chang2024partnr} & $\times$ & $\triangle$ & $\triangle$ & $\triangle$ & $\times$ \\
HandoverSim~\cite{handoversim} & $\triangle$ & $\checkmark$ & $\times$ & $\checkmark$ & $\times$ \\
RoboGen~\cite{wang2023robogen} / Eureka-style generation& $\times$ & $\times$ & $\times$ & $\times$ & $\times$ \\
\textbf{HRIBench} (ours) & $\checkmark$ & $\checkmark$ & $\checkmark$ & $\checkmark$ & $\checkmark$ \\
\bottomrule
\end{tabular}
}
\caption{Qualitative coverage of representative benchmarks. $\checkmark$ indicates explicit support, $\triangle$ indicates partial or indirect coverage, and $\times$ indicates that the axis is not a primary benchmark target.
``Role Script' denotes role-conditioned executable interaction scripts;
``Human Manipulation'' indicates policy-level manipulation under dynamic human motion;
``Perturbation'' refers to perturbation and recovery protocols;
``Interaction Metrics'' denotes interaction-specific evaluation metrics;
and ``Physical Adaptation'' indicates whether physical adaptation studies are provided.
}
\label{tab:benchmark_comparison}
\end{table}

\paragraph{Scope and Task Relevance}
\label{subsec:scope_relevance}

HRIBench prioritizes diagnostic density rather than benchmark scale. 
Each task is designed to expose a specific collaborative failure mode that is difficult to evaluate in isolated manipulation settings, including deictic grounding, handover timing, shared-object synchronization, yielding behavior, and recovery from human intervention. 
As a result, the same manipulation skill may appear under different interaction regimes: a policy capable of grasping an object may still fail when it must wait for a gesture cue, synchronize with a moving collaborator, or stop execution after disruptive human interference.


The role taxonomy is grounded in established HRI requirements. 
\textit{Instructor} tasks evaluate intent communication through demonstrations or gestures~\citep{argall2009survey,villani2018survey}. 
\textit{Collaborator} tasks evaluate close-proximity coordination, where robots must anticipate human motion and execute temporally synchronized actions~\citep{huang2016anticipatory,mainprice2013human,handoversim}. 
\textit{Intruder} tasks evaluate safety and recovery under disruptive human intervention in shared environments~\citep{dragan2013legibility,villani2018survey}. 
This role-based organization keeps the benchmark compact while ensuring that each category corresponds to a distinct coordination challenge.

\paragraph{Scenario Scripts}
\label{subsec:scenario_scripts}

HRIBench represents collaborative interaction through executable \emph{scenario scripts}. 
A script is defined as a sequence of interaction acts:
\begin{equation}
    \mathcal{S} = \{a_1,\ldots,a_T\}, \quad
    a_t = (r_t, g_t, c_t, m_t, \rho_t),
\end{equation}
where $r_t$ denotes the active interaction role, $g_t$ the semantic goal, $c_t$ the temporal or causal constraints, $m_t$ the human motion distribution, and $\rho_t$ the reward and success conditions.

Scenario scripts separate \emph{what} collaboration requires from \emph{how} a policy executes it. 
This abstraction improves interpretability and controllability: scripts can be inspected by humans, compiled into simulation environments, and instantiated with different objects, trajectories, timing offsets, viewpoints, and distractors while preserving the same interaction structure.


\paragraph{Interaction Paradigms}
\label{subsec:paradigms}

\textit{Instructor} scenarios require robots to infer intended actions from human demonstrations or gestures before execution. 
\textit{Collaborator} scenarios require synchronized multi-agent coordination, such as object handover or joint manipulation, where timing and responsiveness are critical. 
\textit{Intruder} scenarios introduce disruptive human interventions and evaluate whether policies maintain safety, reject invalid interference, and recover task execution online. 
Additional details of the Intruder protocol are provided in Appendix F.

\begin{figure}[t]
    \centering
    \includegraphics[
        width=\linewidth,
        trim=34pt 663pt 38pt 41pt,
        clip
    ]{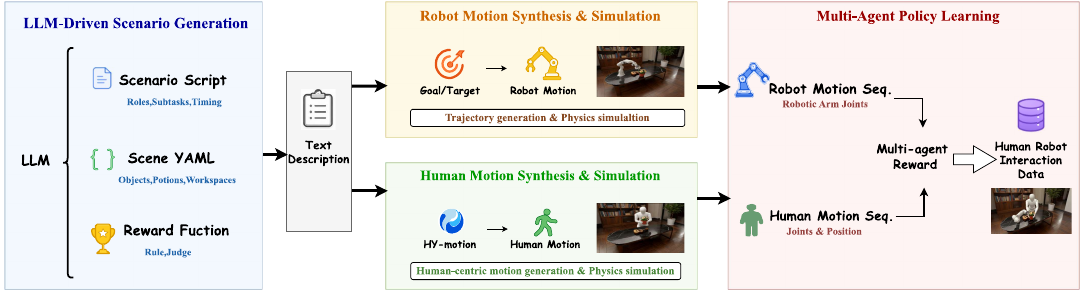}
    \vspace{-14pt}
    \caption{HRIBench generation pipeline. A language model creates scenario scripts, environment specifications, and reward components from task and object metadata. Human and robot trajectories are synthesized, simulated, filtered, and used to construct policy adaptation and evaluation data.}
    \label{fig:pipeline}
\end{figure}

\paragraph{Metrics}
\label{subsec:metrics}
HRIBench evaluates collaboration using three groups of metrics. 
\textbf{Effectiveness} measures task completion quality, including Collaboration Success Rate (CSR), Completion Time (CT), and Idle Ratio (IR). 
\textbf{Coordination} measures interaction timing and protocol adherence, including Temporal Synchronization (TSync), Response Latency (Rsp), and Order Compliance (OC). 
\textbf{Safety and robustness} measures safe and reliable interaction, including Collision-Free Rate (CFR), Human Contact Safety (HCS), Contradictory Instruction Recognition (CIR), and Disruption Success Rate (DSR).
These metrics are reported alongside task success because collaborative failures are often multi-dimensional. 
A policy may complete the manipulation objective while violating temporal coordination, responding to the wrong cue, breaking protocol order, or behaving unsafely around humans. 
Full metric definitions are provided in Appendix A.



\section{Generation Pipeline}
\label{sec:generation_pipeline}

HRIBench converts high-level collaboration specifications into executable interaction trajectories through four stages: scenario scripting, scene instantiation, motion synthesis, and simulation validation. 
The pipeline is designed to preserve the underlying interaction contract while varying scene layout, human behavior, timing, and low-level motion, enabling evaluation of collaborative reasoning rather than memorization of fixed trajectories.


\paragraph{Motion synthesis, rewards, and interaction variants.}
Human interaction prompts are converted into candidate skeleton trajectories using HY-Motion-1.0~\citep{hymotion2025}, followed by task-aware refinement with labeled constraints on hands, torso orientation, and approach directions. 
These structured labels make collaborative behavior directly measurable. 
For example, the benchmark can verify whether a pointing gesture enters the target cue region, whether a receiving hand remains stable during handover, or whether an intruder violates a restricted workspace boundary. HRIBench automatically constructs reward functions and success predicates from task completion, coordination quality, and safety requirements. 
Each task contains more than 50 trajectory variants that preserve the same semantic interaction objective while varying human timing, motion patterns, object poses, and perturbation schedules. 
This diversity prevents overfitting to fixed trajectories and exposes policies to a broad range of coordination conditions. 
Additional generation details are provided in Appendix D.

\paragraph{Simulation validation and filtering.}
All generated scenarios are executed in simulation before inclusion in the benchmark. 
We filter out episodes with unreachable states, severe human-object penetration, invalid camera visibility, inconsistent script ordering, missing success predicates, or labeled joints that fail to satisfy target-region constraints. 
After automatic validation, retained candidates are manually inspected for executability and semantic correctness. 
Overall, 786 out of 2,000 generated interaction trajectories pass the filtering pipeline. 
The resulting validated trajectory pool is then split into policy adaptation and held-out evaluation sets.
For Intruder scenarios, perturbation schedules are generated independently of the evaluated policy and kept fixed across models, ensuring consistent intervention conditions and fair comparison of recovery and safety behavior.
\section{Experiments}
\label{sec:experiments}


The experiments are designed to test the central hypothesis of HRIBench: strong manipulation capability does not necessarily imply robust collaborative intelligence. 
We evaluate representative VLA and imitation-learning policies across the three interaction roles, and further conduct a small-scale real-world adaptation study to examine whether HRIBench-generated interaction data improves physical deployment under limited demonstrations.

\subsection{Implementation Setup}
\label{subsec:implementation_details}
All simulation experiments use a Franka Panda manipulator and are conducted on four NVIDIA A800 GPUs. 
Among 2,000 generated interaction trajectories, 786 pass both automatic validation and manual executable-and-correct screening. 
For each task, 50 validated trajectories are used for policy adaptation and 10 held-out trajectories are reserved for evaluation. 
The evaluation trajectories vary human timing, motion patterns, and scene initialization, preventing policies from simply replaying memorized behaviors.
For the physical adaptation study, we construct task-matched simulation data for a LeRobot SO-100 arm and collect 20 corresponding real-world teleoperation trajectories for each of three representative tasks, one per interaction role.


\subsection{Main Benchmark Results}
\label{subsec:main_results}


We evaluate GR00T N1.5 with LoRA adaptation, $\pi_{0.5}$ with LoRA adaptation, and ACT with full fine-tuning under matched data splits, observation preprocessing, and evaluation seeds whenever supported by the architectures. 
These backbones provide complementary evaluation settings: two pretrained VLA policies with language-conditioned action interfaces, and one imitation-learning baseline based on chunked continuous control. 
Additional backbone details are provided in Appendix G.2. Table~\ref{tab:main_results} reveals a clear role-dependent difficulty progression. 
Instructor scenarios are comparatively easier because the human mainly provides semantic cues, whereas Collaborator scenarios require continuous temporal coordination with a moving partner. 
Intruder scenarios are substantially harder because policies must detect that nominal execution is no longer valid and adapt their behavior online under perturbation.

The VLA-based policies generally achieve higher collaboration success and synchronization scores than ACT, suggesting that pretrained vision-language representations improve cue grounding and interaction-context understanding. 
However, all methods struggle severely in Intruder scenarios. 
Despite perfect or near-perfect Order Compliance (OC), the Collaboration Success Rate (CSR) remains extremely low, indicating that current policies can follow nominal task orderings but fail to properly yield, maintain safe clearance, or recover after disruptive human intervention. 
These results expose a critical gap between manipulation competence and interaction-aware collaborative intelligence. 
Additional analyses are provided in Appendix H.1.

\begin{table}[t]
\centering
\small
\setlength{\tabcolsep}{9pt}
\begin{tabular}{l|l|c|c|c|c|c|c}
\toprule
Role group & Model & CSR$\uparrow$ & CT$\downarrow$ & Rsp$\downarrow$ & TSync$\uparrow$ & OC$\uparrow$ & Safety$\uparrow$ \\
\midrule
\multirow{3}{*}{Instructor}
& GR00T N1.5-LoRA & \textbf{0.533} & 10.5 & \textbf{0.001} & \textbf{0.463} & 0.667 & \textbf{0.993} \\
& $\pi_{0.5}$-LoRA & \textbf{0.533} & \textbf{10.3} & 0.028 & 0.421 & 0.667 & 0.990 \\
& ACT-FT & 0.433 & 14.8 & 0.165 & 0.341 & 0.667 & 0.984 \\
\midrule
\multirow{3}{*}{Collaborator}
& GR00T N1.5-LoRA & \textbf{0.500} & \textbf{18.7} & 0.207 & \textbf{0.394} & 1.000 & \textbf{1.000} \\
& $\pi_{0.5}$-LoRA & 0.457 & 20.1 & 0.146 & 0.355 & 1.000 & 0.994 \\
& ACT-FT & 0.371 & 22.4 & \textbf{0.078} & 0.286 & 1.000 & 0.987 \\
\midrule
\multirow{3}{*}{Intruder}
& GR00T N1.5-LoRA & \textbf{0.100} & \textbf{25.8} & \textbf{0.000} & \textbf{0.276} & 1.000 & \textbf{0.735} \\
& $\pi_{0.5}$-LoRA & \textbf{0.100} & 26.3 & 0.006 & 0.254 & 1.000 & 0.729 \\
& ACT-FT & \textbf{0.100} & 30.0 & 0.016 & 0.224 & 1.000 & 0.721 \\
\bottomrule
\end{tabular}
\caption{Main HRIBench evaluation. Safety is a compact normalized summary of applicable safety and recovery components, including CFR, HCS, CIR, and DSR. Component-wise results are provided in Appendix H.3.}
\label{tab:main_results}
\end{table}
\vspace{-1em}
\subsection{Real-World Adaptation}
\label{subsec:real_world_adaptation}
\vspace{-0.5em}
We further conduct a small-scale real-world adaptation study using GR00T N1.5 on a LeRobot SO-100 arm. 
The objective is not to establish a full physical benchmark, but to evaluate whether HRIBench-generated interaction data provides a useful initialization when real demonstrations are scarce. 
\textbf{Real-only} performs post-training using 20 real teleoperation trajectories per task. 
\textbf{Sim+Real} first adapts the policy using task-matched HRIBench simulation trajectories generated for SO-100, followed by post-training on the same real-world demonstrations. 
We report task success rate over 10 evaluation trials per task.

\begin{table}[t]
\centering
\small
\setlength{\tabcolsep}{9pt}
\renewcommand{\arraystretch}{1}
\begin{tabular}{l|l|c|c}
\toprule
Role group & Real-world task & Real-only SR$\uparrow$ & Sim+Real SR$\uparrow$ \\
\midrule
Instructor 
& Deliver via gesture commands 
& 2/10 & \textbf{5/10} \\
Collaborator 
& Collaborative object retrieval from a box 
& 1/10 & \textbf{4/10} \\
Intruder 
& Avoid having someone else do the placement 
& 0/10 & \textbf{4/10} \\
\midrule
Average 
& --
& 0.10 & \textbf{0.43} \\
\bottomrule
\end{tabular}
\caption{Real-world adaptation comparison with GR00T N1.5 on a LeRobot SO-100 arm.}
\label{tab:real_world_adaptation}
\end{table}

Across 30 physical evaluation trials, Real-only succeeds in only 3 cases, while Sim+Real succeeds in 13 cases. 
The improvement is consistent across all three interaction roles, including gesture-conditioned execution, collaborative retrieval in shared workspaces, and interference-aware placement. 
Although the physical study remains limited in scale, the results suggest that structured interaction data from HRIBench provides transferable priors for real-world collaborative behavior.


\begin{figure}[t]
    \centering
    \includegraphics[
        width=0.9\linewidth,
        height=0.45\textheight,
        trim={1.2cm 19cm 8.32cm 1.5cm},
        clip
    ]{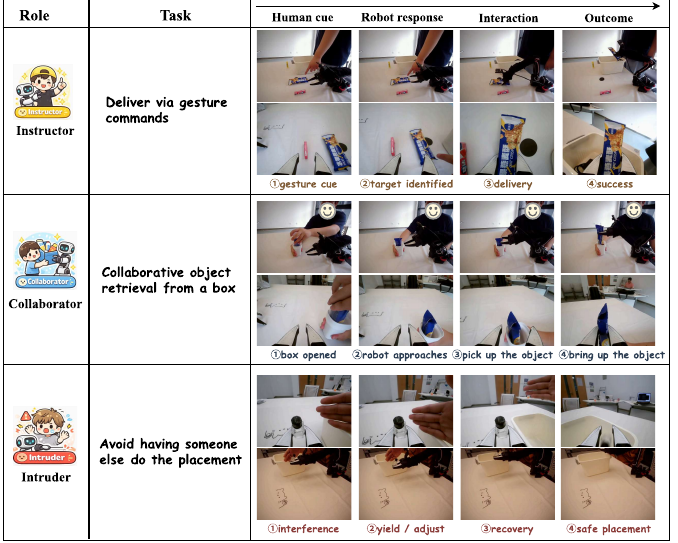}
    \vspace{-8pt}
    \caption{Representative real-world SO-100 tasks: gesture-conditioned delivery, collaborative retrieval, and interference-aware placement.}
    \label{fig:real_world_demo}
\end{figure}

\vspace{-1em}
\subsection{Failure Mode Analysis}
\label{subsec:failure_mode_analysis}
\vspace{-0.5em}

Beyond aggregate success rates, HRIBench exposes distinct role-dependent collaborative failure modes. 
Instructor failures are primarily caused by incorrect cue grounding or premature execution before the human signal is completed. 
Collaborator failures often occur despite correct manipulation intent, because the robot reacts too early, too late, or fails to synchronize with the human partner during shared-object interaction. 
Intruder failures are qualitatively different: policies frequently continue the original execution plan after disruptive human intervention, leading to unsafe proximity, invalid recovery behavior, or failure to restore the intended task state.
These findings highlight why binary task success alone is insufficient for evaluating collaborative embodied intelligence. 
A policy may complete the object-level goal while still violating temporal coordination, interaction protocol, or safety requirements. 
The full failure-mode breakdown is provided in Appendix H.1.

\begin{figure}[t]
    \centering
    \includegraphics[
        width=0.9\linewidth,
        height=\textheight,
        keepaspectratio,
        trim=24pt 176pt 10pt 93pt,
        clip
    ]{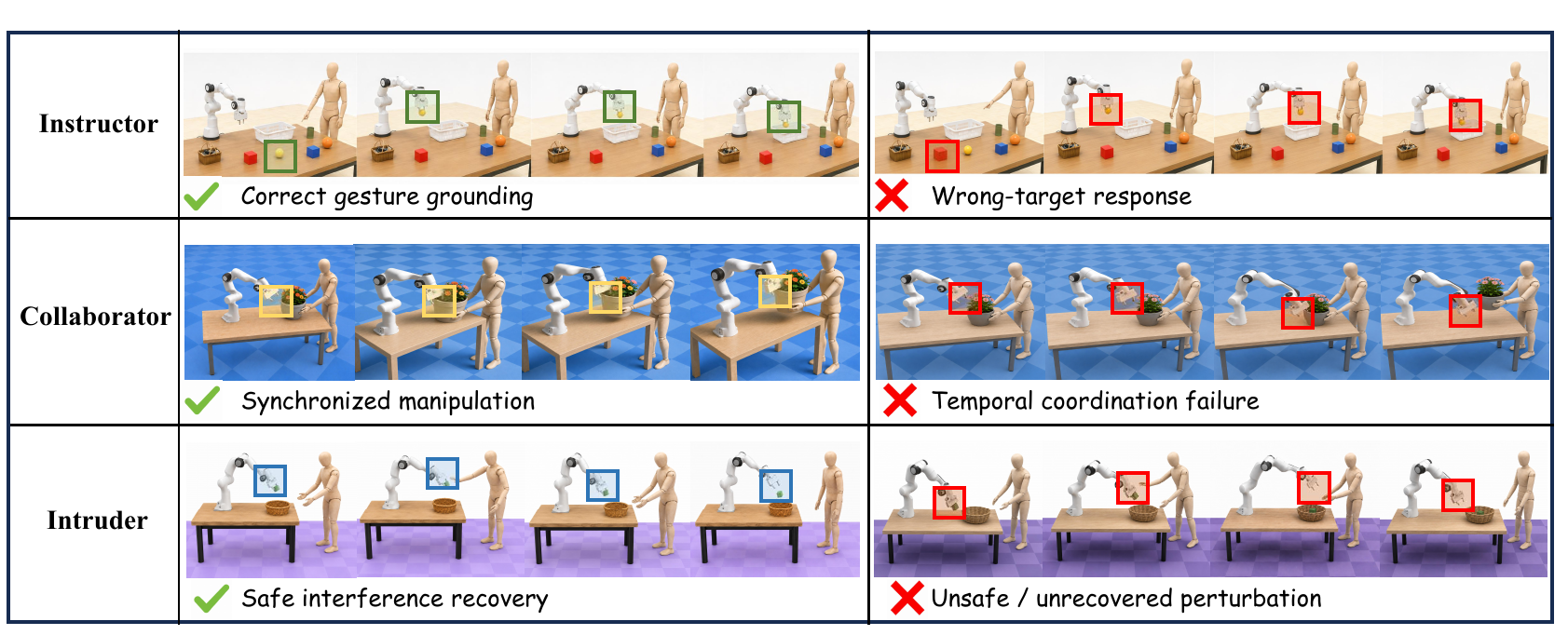}
       \vspace{-8pt}
    \caption{Representative successful and failed HRIBench rollouts across different interaction roles.}
    \label{fig:success_failure_results}
\end{figure}

\vspace{-1.5em}
\section{Conclusion}
\label{sec:conclusion}

\vspace{-1em}
HRIBench reframes human--robot collaboration as executable, role-conditioned interaction rather than isolated manipulation. By combining scenario scripts, generated human motion, simulation validation, and interaction-centric metrics, it diagnoses collaborative failures in terms of intent grounding, temporal coordination, safety, and recovery. Experiments with GR00T N1.5, $\pi_{0.5}$, and ACT show that current robot policies remain substantially limited in collaborative settings, especially under human intervention. A small-scale SO-100 study further suggests that HRIBench-generated simulation data can improve real-world adaptation when demonstrations are scarce. Overall, HRIBench provides a diagnostic benchmark for evaluating and improving interaction-aware robot learning.
\vspace{-1.2em}

\section{Limitations}
\label{sec:limitations}

\vspace{-1.1em}
HRIBench is a simulated benchmark and should not be treated as a real-world safety evaluation. Although human motion is generated from HY-Motion-1.0 candidates and refined with labeled skeleton objectives, simulated avatars cannot fully capture the variability and ambiguity of real human behavior. Contact is approximated through hand labels and object-proximity constraints rather than high-fidelity physical interaction, so safety metrics should be interpreted together with clearance and event-level behaviors.
Our SO-100 experiments are small-scale and intended to evaluate sim-to-real transfer rather than deployment robustness. In addition, differences in pretraining data, embodiment assumptions, and adaptation protocols may affect direct comparison across robot policies.


\clearpage
\bibliography{references}
\clearpage

\end{document}